\documentclass[preprint,12pt,authoryear]{elsarticle}
\usepackage{amsmath,amsfonts}
\usepackage{algorithmic}
\usepackage{algorithm}
\usepackage{array}
\usepackage[caption=false,font=normalsize,labelfont=sf,textfont=sf]{subfig}
\usepackage{textcomp}
\usepackage{stfloats}
\usepackage{url}
\usepackage{verbatim}
\usepackage{graphicx}
\usepackage{natbib}

\usepackage{booktabs}
\usepackage{tablefootnote}
\usepackage{newfloat}
\usepackage{listings}
\usepackage{algorithm}
\usepackage{algorithmic}
\usepackage{amssymb}
\usepackage{multirow}
\usepackage{xcolor,colortbl}
\usepackage{kotex}
\usepackage{makecell}
\newcommand{\ie}{\textit{i}.\textit{e}., }
\newcommand{\eg}{\textit{e}.\textit{g}., }

\newcommand{\etal}{\textit{et al}.}

\journal{Expert Systems with Applications}
\begin{document}
\begin{frontmatter}
\title{Spatial Bias for Attention-free Non-local Neural Networks}

\author[ajou2]{Junhyung Go}
\author[ajou1,ajou2]{Jongbin Ryu}
\address[ajou1]{Department of Computer Engineering, Ajou University}
\address[ajou2]{Department of Artificial Intelligence, Ajou University}

\vspace{-2.0em}
\begin{abstract}
In this paper, we introduce the spatial bias to learn global knowledge without self-attention in convolutional neural networks.
Owing to the limited receptive field, conventional convolutional neural networks suffer from learning long-range dependencies. 
Non-local neural networks have struggled to learn global knowledge, but unavoidably have too heavy a network design due to the self-attention operation.
Therefore, we propose a fast and lightweight spatial bias that efficiently encodes global knowledge without self-attention on convolutional neural networks.
Spatial bias is stacked on the feature map and convolved together to adjust the spatial structure of the convolutional features. 
Therefore, we learn the global knowledge on the convolution layer directly with very few additional resources.
Our method is very fast and lightweight due to the attention-free non-local method while improving the performance of neural networks considerably. 
Compared to non-local neural networks, the spatial bias use about $\times 10$ times fewer parameters while achieving comparable performance with $1.6\sim3.3$ times more throughput on a very little budget.
Furthermore, the spatial bias can be used with conventional non-local neural networks to further improve the performance of the backbone model.
We show that the spatial bias achieves competitive performance that improves the classification accuracy by $+0.79\%$ and $+1.5\%$ on ImageNet-1K and cifar100 datasets.
Additionally, we validate our method on the MS-COCO and ADE20K datasets for downstream tasks involving object detection and semantic segmentation.
\end{abstract}


\end{frontmatter}
\section{Introduction}
\begin{figure}[b!]
\centering
\includegraphics[width=6.8cm, height=6.62cm]{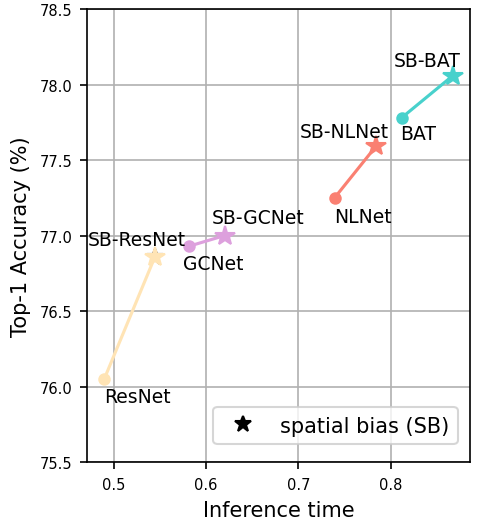}
\caption{Performance comparison for the proposed method and conventional non-local neural networks on ImageNet-1K dataset. {\bf $\bullet$} denotes the naive ResNet backbone and conventional non-local neural networks and {\bf $\star$} present performance improvement of the conventional networks with our spatial bias. 
In all cases, the proposed spatial bias enhance networks with minimal computational complexity.
}
\label{figure:acc_throughput}
\end{figure}

Convolutional neural networks (CNNs) excel in extracting nuanced local information.
Thanks to this advantage, CNNs are utilized for a variety of visual recognition tasks.
However, their inability to effectively capture the global context has been mentioned numerous repeatedly.
Due to the limited receptive field size, the convolution focuses on a small region that learns only local information; to overcome this, several approaches to increase the receptive field size have been extensively studied, such as building deeper layers~\cite{resnet}, employing different kernel sizes~\cite{googlenet,sknet,li2022faster}, and learning non-local pixel-level pairwise relationships~\cite{nln,gcnet,spanet,you2022nonlocal,expert_nonlocal_fractal,cho2022rethinking,nlnbilinear,expert_nonlocal_robust}.
Among these methods, self-attention based non-local neural networks~\cite{nln} have been a major approach to capture long-range information.
However, they exploit an excessive amount of resources because of the self-attention operation.
Therefore, this paper presents a novel lightweight method that directly learns the long-range dependency during the convolution operation.
The proposed method acquires global knowledge by incorporating a spatial bias term into the convolution operation.
A spatial bias with long-range correlation is added to the position in which convolution is performed.

The proposed method allows for the simultaneous learning of local information from the convolution term and global knowledge from the spatial bias term.
In addition, a minimal amount of resources are used for the proposed spatial bias term, and thus our method is very fast and lightweight compared to the conventional self-attention based non-local neural networks.
We extensively carry out experiments on the number of parameters, FLOPs, throughput, and accuracy to show the efficacy of the proposed spatial bias method.
As shown in Fig.~\ref{figure:acc_throughput}, the inference time overhead of our spatial bias is {\bf 1.6} to {\bf 3.3} times less than that of non-local neural network while achieving competitive performance compared to the non-local networks~\cite{nln,gcnet,nlnbilinear}.
The proposed spatial bias further improves the performance of backbone networks in conjunction with existing self-attention operations.
The following is a summary of the contributions regarding our spatial bias.

\begin{itemize}
    \item 
    We introduce a spatial bias that takes into account both local and global knowledge in a convolution operation.
    Thanks to the proposed lightweight architecture, the spatial bias term is computed very quickly and with a small amount of overhead.
    \item 
    We show that the proposed spatial bias term significantly improves the performance while incurring very modest overheads: in the case of ResNet-50 backbone, the parameter overhead of spatial bias has only $6.4\%$ and $\times3.3$ faster compared to the non-local neural network (NLNet)~\cite{nln}.
    \item 
    We verify the generalizability of the proposed spatial bias by combining it with other non-local methods and backbones. We also confirm that spatial bias improves performance in downstream tasks.

\end{itemize}

\begin{figure}[t]
\centering
\includegraphics[width=0.6\textwidth]{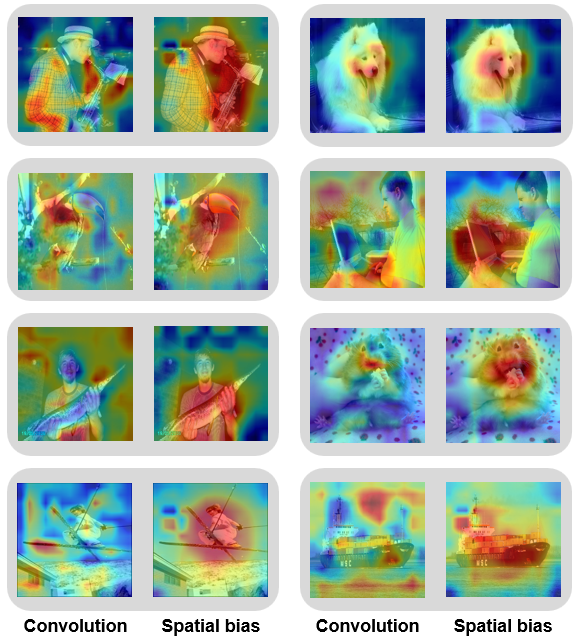} 
\caption{Grad-Cam~\cite{gradcam} visualization for spatial bias and convolution feature map. 
Notably, the grad-cam on the spatial bias exposes wider regions, which aids in learning global knowledge.
}
\label{spatial_bias_img}
\end{figure}

\section{Related Work}

\paragraph{Non-local Neural Network with Self-attention}

The non-local neural networks~\cite{nln, gcnet, nlnbilinear} using self-attention operation that learns long-range dependency has been most widely studied.
Unlike convolution operation, self-attention learns global knowledge in a single layer, which alleviates the narrow receptive field problem of CNNs.
This approach performs well when applied to a variety of visual tasks.
In particular, NLNet~\cite{nln} is the first study to exploit the self-attention operation for learning the pairwise relationship at the global pixel-level.
However, since NLNet is calculated after obtaining an attention map independent of each query, the computation complexity is very high. 
For different query locations, GCNet~\cite{gcnet} generates similar attention maps, thereby modeling an efficient non-local block. 
In order to create lighter non-local networks, CCNet~\cite{ccnet}, A2Net~\cite{a2net}, and BAT~\cite{nlnbilinear} have been introduced by using an iterative cross-attention module~\cite{ccnet}, dual-attention method~\cite{a2net}, and data-adaptive bilinear attentional transformation~\cite{nlnbilinear}.
Fang \etal \cite{spanet} proposes an location-based attention method that distills positive information in a global context.
Recently, the non-local neural networks are used to various tasks such as histopathological image classification~\cite{expert_nonlocal_fractal} and hand detection~\cite{expert_nonlocal_robust}.
These methods have contributed to the design paradigm of non-local neural networks with the reduced overhead of self-attention operation.
However, we argue that they still suffer from a fatal flaw in that their computational cost is quadratic {\bf \(O(n^{2})\)}\footnote{$n$ indicates the number of all positions in the feature map} which causes a slowdown of inference time.

Due to the heavy design of self-attention, conventional non-local methods are inserted only at specific layers in a convolutional neural network by consideration of the throughput and network size.
Additionally, since traditional non-local operations only consider spatial correlation by merging channels, they are blind to any channel correlation.
Therefore, to overcome these limitations, we propose spatial bias, an attention-free non-local neural network with a fast and lightweight architecture.
In comparison to self-attention based non-local neural networks, the proposed spatial bias achieves comparable performance with $1.6\sim3.3$ times more throughput on a very little budget.
Additionally, lightweight spatial bias can be employed across all network levels, complementing the existing heavy self-attention that can be given to particular layers.
Thus, our spatial bias enables a network to learn more about global knowledge due to its fast and lightweight properties.

\paragraph{Architecture Modeling}

Recently, effective neural networks have shown advances across a range of many visual recognition tasks.
The modern CNN architecture conceived by LeNet~\cite{lenet} was realized a decade ago by AlexNet~\cite{alexnet}, and various studies have been conducted to improve its performance and usefulness. 
Since then, CNN~\cite{vgg} with small filter sizes has been developed to encode more precise regional data.
Consequently, skip connections have made it possible to build deeper networks~\cite{resnet} and several studies have been done to increase expressiveness by varying the width, multi-path block design, or grouped feature flow of neural networks~\cite{wideresnet,googlenet,resnext,expert_gao2023very_group,schwarz2022enhanced}.
Through a multi-branch design, recent CNN architectures~\cite{googlenet, sknet, resnest, res2net,guo2021multi} communicate information between branches.
Additionally, several methods~\cite{nln,nlnbilinear,gcnet} capture long-range dependencies by taking advantage of the self-attention operation that guarantees a better understanding of global knowledge for the visual recognition task.

\begin{figure*}[t]
\centering
\includegraphics[width=1\textwidth, height=4.5cm]{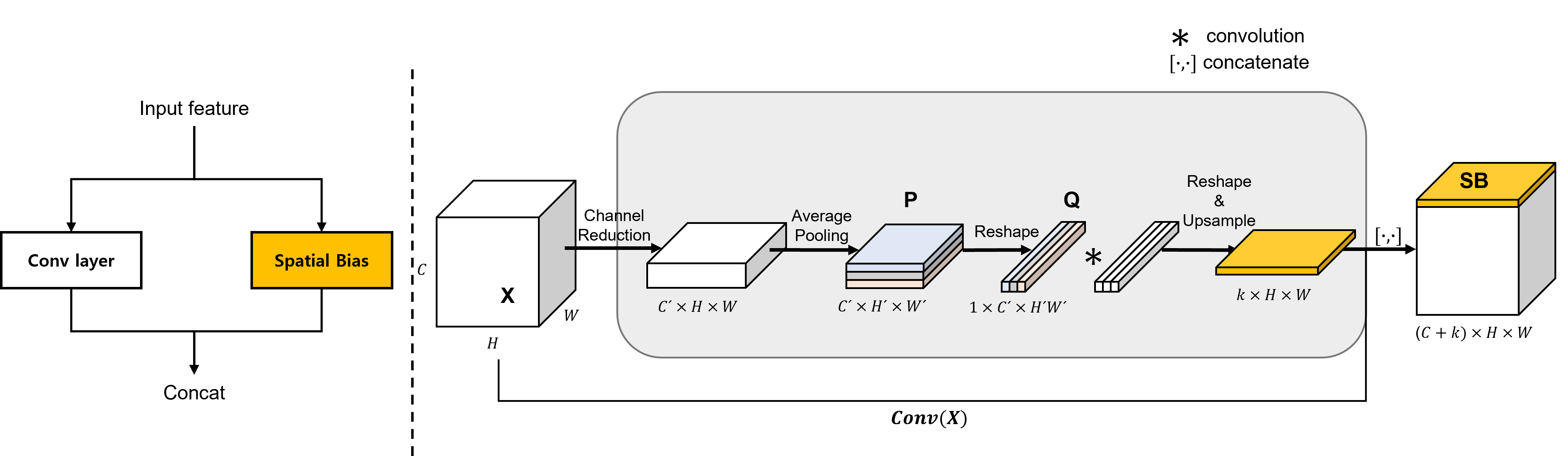} 
\caption{Design of Spatial Bias term. 
The figure on the left shows the overall workflow and the right one shows its detail. In the right figure, to capture the global dependency, the channel and spatial size of the feature map are reduced through the $1\times1$ convolution and average pooling operations. We aggregate spatial bias on a reduced feature map using a simple 1D convolution operation.}
\label{figure_overflow}
\end{figure*}
\section{Proposed Method}

The convolution utilizes a shared weight within a limited local window, allowing CNN to have the property of translational equivalence~\cite{translation_equi}.
Recently, this property has been identified as the inductive bias~\cite{inductive_bias}, and it has been stated repeatedly that convolution is not particularly effective at learning the relationship between long-range objects~\cite{nln}.
To address this problem, we propose a spatial bias term to compensate for these shortcomings in the convolution.
Different from the existing method~\cite{nln, gcnet, nlnbilinear} using the heavy self-attention module, the proposed method learns the global knowledge by adding a few spatial bias channels to the convolution feature map.
Inspired by parallel network designs~\cite{googlenet,sknet,resnest,res2net}, we devise the parallel branches that could complement long-range dependencies of a network as shown in Fig.~\ref{figure_overflow}. 
To generate the spatial bias, we encode long-range dependencies by compressing feature maps in channel and spatial directions.
Then, we extend it to concatenate the spatial bias map and the convolutional feature map in the channel direction.
Global knowledge from spatial bias is aggregated with the local features of the convolutional feature map, so the network learns both local and global information. 
As shown in the Fig.~\ref{spatial_bias_img}, the spatial bias learns a wider region while convolution focuses on the local features in an image.
Therefore, the CNN with the spatial bias learns richer information through our concatenated feature map.
The following section introduces the specific process of aggregating convolutional feature map and spatial bias.

\paragraph{Generating Spatial Bias Map}

Let the input feature map $X$ of a convolution layer be \(X\in\mathbb{R}^{H\times{W}\times{C}}\).
On this feature map, we compress it in the channel and spatial direction. 
Specifically, we use $1\times1$ convolution for channel reduction, where the feature map has a $C^{\prime}$ channel.
Then, we adopt an average pooling layer for the spatial compression that yields \(P\in\mathbb{R}^{H^{\prime}\times{W^{\prime}}\times{C^{\prime}}}\).
We get a transformed feature map by flattening each channel of the feature map P into a 1D vector,
\(Q\in\mathbb{R}^{{1}\times{C^\prime}\times{H^{\prime}{W^{\prime}}}}\).

To aggregate global knowledge on this transformed feature map, we exploit a $1\times N$ convolution in the channel dimension where the $N$ is larger than $1$ so that we encode the inter-channel relationship global knowledge.
The spatial bias map is then upsampled to the same size as the convolutional feature map using bilinear interpolation.
The upsampled spatial bias map is concatenated with the convolutional feature maps as Eq.~\ref{eq:sb_layer}.
\begin{equation}
    Output = ReLU(BN[Conv(X),SB]),
\label{eq:sb_layer}
\end{equation}
where $X$ denotes an input feature map, $Conv()$ and $SB$ denote a standard convolution and a spatial bias, and [,] is the concatenate operation. 
After concatenation, the resultant feature map is sent through batch normalization and nonlinear activation layers.

\paragraph{Convolution with Spatial Bias}

In general, naive convolution employs modest kernel sizes (\eg $3 \times 3$). 
To compensate the limited kernel size, the self-attention module is added after specific convolution layer to learn the global knowledge.
In other words, the heavy self-attention operation is independently applied which increase parameters and computational budget extensively. 
The proposed spatial bias and convolutional features are complementary to each other.
Spatial bias extracts the information of long range-dependency, which complement the existing short range-dependency of convolutional operation.
The proposed spatially biased convolution need only minimal overhead of convolution operation due to our self-attention free approach.
We aim to learn both of the local and global knowledge in convolution layer directly without additional module.

\paragraph{Complexity of Spatial Bias}
In this paragraph, we discuss about the complexity of the proposed spatial bias in comparison with the self-attention operation.
Suppose input size is defined as \(X\in\mathbb{R}^{H\times{W}\times{C}}\), the self-attention mechanism has a computational complexity of $O((HW)^{2}C)\approx{O((HW)^{2})}$, because self-attention operation applies three projections to obtain query, keys and values and computes dot products between query and keys.
On the other hand, the proposed spatial bias reduces the feature map size $X$ by a fixed constant.
Therefore, the complexity of spatial bias is $O(H^{\prime}W^{\prime}sf)\approx{O(H^{\prime}W^{\prime}f)}$,
where $s$ and $f$ denote the kernel size, number of filters.
Since the number of filters is the same as $H^{\prime}W^{\prime}$, the spatial bias has the complexity of ${O((H^{\prime}W^{\prime})^{2})}$.
The reduced height $H^{\prime}$ and width ${W^{\prime}}$ are fixed constant value, so the computational complexity is ideally ${O(1)}$.

Therefore, we get very fast and lightweight operation that can be inserted into any convolutional layers.
In the experiment section, we show its effectiveness with regard to the throughput, parameters, and computational overhead as well as performance improvement of CNNs.

\paragraph{On the Comparison with Squeeze and Excitation}
The general channel attention method (\ie SENet, SKNet)~\cite{senet,sknet} captures the channel-wise summarized information of the feature map and then learns the relationship between the channels to adjust the feature map, so the spatial correlation is not learned.
On the other hand, the proposed spatial bias extracts the spatial-wise compressed information and then expands it toward the channel.
That is, dependence on the spatial-channel direction can be aggregated at once with only a general convolution operation.
In addition, while SE-like method refine the channel importance of the output feature map of the convolution layer, the proposed method learn different information in that it learns the channel and spatial information together in the convolution process directly by adding a bias to the feature map.
Therefore, as shown in Table~\ref{imagenet_results:attention}, the proposed spatial bias is more efficient than SE-like methods and additionally improve the performance of backbone when combining them with our spatial bias.

\section{Experiments}
\label{sec:exp}

\begin{table}[t]
\renewcommand{\arraystretch}{1.2}
\setlength{\tabcolsep}{11pt}
\centering
\caption{Experimental result on CIFAR-100 through the proposed Spatial Bias({\bf 3-bias channels}). Here, $s_{\#}$ denotes the stage index of the ResNet architecture after the stem cell.}
{\scriptsize
\begin{tabular}{l|l|cccc}
\toprule
\rowcolor{gray!30}
Network & Stage & Param. & MFLOPs & Top-1 Error (\%) & Top-5 Error (\%) \\
\midrule
\multirow{4}{*}{ResNet-38} & -     & 0.43M & 62.2 & 23.98 $\pm 0.23$ & 5.73 $\pm 0.20$ \\
                          & $s_{1}$       & 0.45M & 63.6 & 23.86 $\pm 0.08$ & 5.60 $\pm 0.13$ \\
                          & \cellcolor{red!20}$s_{1},s_{2}$  & \cellcolor{red!20}0.46M & \cellcolor{red!20}64.2 & \cellcolor{red!20}22.44 $\pm 0.10$& \cellcolor{red!20}4.99 $\pm 0.05$ \\
                          & $s_{1},s_{2},s_{3}$ & 0.48M & 64.6 & 22.46 $\pm 0.25$& 5.21 $\pm 0.30$ \\
\midrule
\multirow{4}{*}{ResNet-65} & -     & 0.71M & 103.3 & 21.87 $\pm 0.35$& 5.33 $\pm 0.18$\\
                      & $s_{1}$       & 0.74M & 105.8 & 20.81 $\pm 0.41$& 4.73 $\pm 0.09$\\
                      &\cellcolor{red!20}$s_{1},s_{2}$    & \cellcolor{red!20}0.77M & \cellcolor{red!20}106.9 & \cellcolor{red!20}20.77 $\pm 0.14$& \cellcolor{red!20}4.67 $\pm 0.29$\\
                      & $s_{1},s_{2},s_{3}$ & 0.80M  & 107.5 & 20.37 $\pm 0.20$& 4.61 $\pm 0.09$\\
\midrule
\multirow{4}{*}{ResNet-110} & -     & 1.17M & 171.9 & 20.59 $\pm 0.38$& 4.96 $\pm 0.10$\\
                        & $s_{1}$       & 1.22M & 176.1 & 19.97 $\pm 0.08$& 4.55 $\pm 0.06$\\
                        &\cellcolor{red!20}$s_{1},s_{2}$    & \cellcolor{red!20}1.28M & \cellcolor{red!20}178.0 & \cellcolor{red!20}19.42 $\pm 0.20$& \cellcolor{red!20}4.38 $\pm 0.06$\\
                        & $s_{1},s_{2},s_{3}$ & 1.34M & 179.1 & 19.65 $\pm 0.68$& 4.80 $\pm 0.21$\\
\bottomrule
\end{tabular}
}
\label{table_cifar_depth}
\end{table}

\begin{table}[t]
\renewcommand{\arraystretch}{1.2}
\setlength{\tabcolsep}{13pt}
\centering
\caption{Experimental result on the comparison of insertion position in a bottleneck. We add spatial bias in parallel after {\bf Conv1} or {\bf Conv2}. The out channels of {\bf Conv2} is reduced so that the spatial bias after {\bf Conv2} has less parameters.
}
{\scriptsize
\begin{tabular}{llcc}
\toprule
\rowcolor{gray!30}
Position  & Stage & Top-1 Error (\%) & Param. \\
\midrule
 \multirow{2}{*}{\bf Conv1} & $s_{1},s_{2}$    & 20.57    & 0.78M  \\
                              & $s_{1},s_{2},s_{3}$ & 20.56 & 0.83M  \\
\midrule
 \multirow{2}{*}{\bf Conv2} & $s_{1},s_{2}$    & 20.77    & 0.77M  \\
                              & $s_{1},s_{2},s_{3}$ & 20.37 & 0.80M \\
\bottomrule
\end{tabular}
}
\label{table_insert_position}
\end{table}

In this section, we first perform ablation studies on the proposed spatial bias, then compare it with the conventional non-local neural networks with self-attention operation. 
We, then, provide the experimental analysis of OOD and shape bias to support the effectiveness of the spatial bias.   
Finally, we show the experimental result on the object detection and semantic segmentation tasks. 

\subsection{Experimental result on CIFAR-100}

\paragraph{Setup}
We report mean accuracy of three experiments using the CIFAR-100 dataset on the proposed method with different random seed for reliable comparison.
We set the training recipe with reference to~\cite{cutmix}. 
We use $32\times32$ image size 64 samples per mini-batch with 300 epochs.
We initially set the learning rate as 0.25 and decayed it by a factor of 0.1 after each 75 epohs.

\paragraph{Position of Spatial Bias}\label{section:location}
Table ~\ref{table_cifar_depth} summarizes the performance comparison of spatial bias positions in ResNet stages.
Since the spatial bias compresses the spatial resolution to the fixed size (\ie{6 for CIFAR-100, 10 for ImageNet}, Table~\ref{new_cifar_reso}), the overhead of parameters and computational budget is very small regardless of the stages.
When we apply the spatial bias to stage1 ($s1$) and stage2 ($s2$), the performance of ResNet backbone is improved considerably. 
However, in the last stage ($s3$), there is no performance improvement with the spatial bias. 
We assume that the spatial size of the last stage is too small so that the global knowledge is disappeared.

\begin{table}[t]
\renewcommand{\arraystretch}{1.2}
\setlength{\tabcolsep}{12pt}
\centering
\caption{Various size for Spatial bias on CIFAR-100 (\ie $SB_{6}$ denotes compression size of 6).}
{\scriptsize
\begin{tabular}{lccc}
\toprule
\rowcolor{gray!30}
Network & Param. & Top-1 Error (\%) & \makecell[c]{Throughput \\  (sample/sec)}\\
\midrule
ResNet-65 & 0.71M & 21.87 & 12,816\\
$SB_{6}$ & 0.77M & 20.77 & 10,276 \\
$SB_{10}$ & 1.13M & 20.48 & 10,221\\
$SB_{14}$ & 2.33M & 20.84 & 10,267 \\
$SB_{16}$ & 3.47M & 20.75 & 10,467\\
\bottomrule
\end{tabular}
}
\label{new_cifar_reso}
\end{table}

Further, we validate the position of the spatial bias in a residual bottleneck. 
Table~\ref{table_insert_position} shows the performance comparison in the insertion position of spatial bias after {\bf Conv1} or {\bf Conv2} in a bottleneck. 
We confirm that the spatial bias after {\bf Conv2} reduce the parameters while achieving similar performance compared to that of {\bf Conv1}.

\paragraph{Number of Spatial Bias Channels}
We compare the performance on the number of spatial bias channels.
The channel of the spatial bias represents its importance in the concatenated output, and thus the more channels are used, the more global knowledge will be learned from the spatial bias.
As shown in Table~\ref{table_number_of_bias}, we confirm that the performance and overhead are the optimal when three channels were used ({\bf Bias-3} and {\bf Bias-4}), but the performance is degraded beyond that (\ie {\bf Bias-5} and {\bf Bias-6}).
It is inferred that if too much spatial bias is used, the convolution features are damaged which cause the performance deteriorates of entire networks.
\begin{table}[t]
\renewcommand{\arraystretch}{1.2}
\setlength{\tabcolsep}{23pt}
\centering
\small
\caption{Experimental results on the number of spatial bias channels in CIFAR-100. Bias-\# indicates the number of channels.
We use ResNet-65 as the backbone networks. We add the spatial bias to the stage 1 and 2.
}
{\scriptsize
\begin{tabular}{l|cc}
\toprule
\rowcolor{gray!30}
Method & Param. & Top-1 Error (\%) \\
\midrule
Bias-0 & 0.71M & 21.87  \\

\midrule
Bias-1   & 0.76M &  20.91  \\
\midrule
Bias-2 & 0.77M &  20.99  \\
\midrule
Bias-3  & 0.77M  & {\bf 20.77}  \\
\midrule
Bias-4  &  0.77M & {\bf 20.60}    \\
\midrule
Bias-5 & 0.77M & 21.06\\ 
\midrule
Bias-6 & 0.77M & 20.82\\
\bottomrule
\end{tabular}
}
\label{table_number_of_bias}
\end{table}
\paragraph{Component analysis}
We perform an ablation study on spatial bias component analysis. {\bf Add} in Table ~\ref{new_component} represent that the spatial bias is added to the feature map. In addition, the average pooling layer is replaced by the maximum pooling layer ({\bf Maxpool}). Lastly, the global context is aggregated with only the average layer({\bf Pool only}) for performance verification on the key operations.

\begin{table}
\renewcommand{\arraystretch}{1.2}
\setlength{\tabcolsep}{24pt}
\centering
\caption{Component analysis for Spatial bias on CIFAR-100.}
\scriptsize{
\begin{tabular}{lcc}
\toprule
\rowcolor{gray!30}
Network & Param. & Top-1 Error (\%)\\
\midrule
Add & 0.76M & 20.93  \\
Maxpool & 0.77M & 21.03 \\
Pool only & 0.71M & 22.34 \\
\midrule
$SB_{6}$ & 0.77M & 20.77 \\
\bottomrule
\end{tabular}
}
\label{new_component}
\end{table}

\begin{table}[t]
\renewcommand{\arraystretch}{1.2}
\setlength{\tabcolsep}{17pt}
\centering
\caption{Experimental results on the position of spatial bias in ImageNet-1K.}
\scriptsize{
\begin{tabular}{l|lccc}
\toprule
\rowcolor{gray!30}
Method & Stage & Param. & Top-1 (\%) & Top-5 (\%)\\
\midrule
Bias-0 & -    & 25.56M &  76.42 & 92.87\\
\midrule
\multirow{4}{*}{Bias-3} & $s_{1},s_{3}$    &   25.86M &  76.68       & 93.13\\
 & $s_{2},s_{3}$    &   25.89M &  77.00       & 93.00\\
 & \cellcolor{red!20}$s_{1},s_{2},s_{3}$    &   \cellcolor{red!20}25.99M &  \cellcolor{red!20}{\bf 77.11} & \cellcolor{red!20}93.19\\
 & $s_{1},s_{2},s_{3},s_{4}$    &   26.02M &  76.70       & 93.08\\
\bottomrule
\end{tabular}
}
\label{imagenet:table_spatial_ratio}
\end{table}

\subsection{Experimental results on ImageNet}
\paragraph{Setup}
In this section, we present experimental result on ImageNet-1k, which includes more than 1.2 million images with 1,000 class labels.
We validate our performance on ImageNet dataset using two training recipes. 
First, we use the training recipe of~\cite{resnet-strike} to demonstrates the effectiveness of the proposed spatial bias.
In this recipe, the training mini-batch size is set to 512 with 100 epochs using $160\times160$ input size.
We initialize the learning rate to 8e-3 with the cosine learning rate scheduler.
For optimization, we use a lamb~\cite{lamb} that is suitable for training of large batch size.
Second, we follow NLNet~\cite{nlnbilinear}'s training recipe for fair comparison with state-of-the-art non-local neural networks.  
Specifically, we exploit the cropped input image as $224\times224$ size.
The initial learning rate of 0.1 is reduced by 0.1 after 30, 60, and 80 epochs.
We use 256 batch size and stochastic gradient descent(SGD) optimizer.

\paragraph{Result} 
We perform experiments with the position of the spatial bias and its channel size. 
We compare the performance of spatial bias on which stage of ResNet backbone.
As shown in Table~\ref{imagenet:table_spatial_ratio}, the spatial bias has the best performance when adding it from stage 1 to 3 \(s1, s2, s3\).
This result is the same as that of CIFAR-100 result where the spatial bias does not work on the small spatial size of the last stage.
In addition, when the spatial bias is not used at the first stage \(s1\), the performance increase is insignificant. 
This result means that, as in previous studies~\cite{nln,nlnbilinear}, global knowledge exists a lot in the earlier layer with high resolution, and thus the effect of spatial bias is greater.

Table~\ref{imagenet:number_of_bias} shows the performance comparison in the number of spatial bias channels in ImageNet dataset.
When $3$$\sim$$4$ channels of the spatial bias are added, the performance is improved by $+0.58\%$ compared to the baseline, and wider channels $5$$\sim$$6$ degrade the performance.
This result also confirm that the optimal channels should be used to avoid the damage of convolution feature map.

\begin{table}[t]
\renewcommand{\arraystretch}{1.2}
\setlength{\tabcolsep}{14pt}
\centering
\caption{Experimental result on the number of spatial bias channels in ImageNet-1K.} 
\scriptsize{
\begin{tabular}{l|cc>{\columncolor{red!20}}c>{\columncolor{red!20}}cc}
\toprule
\rowcolor{gray!30}
Bias-\# & 1 & 2 & 3 & 4 & 5 \\
\midrule
Top-1 (\%) & 76.70 & 76.84 & {\bf 77.00} & {\bf 77.00} & 76.66 \\
\midrule
Param. & 25.87M & 25.88M & {\bf 25.89M} & {\bf 25.90M} & 25.91M \\
\bottomrule
\end{tabular}
}
\label{imagenet:number_of_bias}
\end{table}

\begin{table}[t]
\renewcommand{\arraystretch}{1.2}
\setlength{\tabcolsep}{7.1pt}
\centering
\caption{Experimental results on the standard attention operation. Unlike channel attention operation, spatial bias learns channel and spatial-wise dependencies to readjust the feature map. 
In addition, our spatial bias is lighter than channel attention operation, and has faster inference speed. Therefore, the channel attention network to which spatial bias is added improves performance with only a very small additional budget.}
{\scriptsize
\begin{tabular}{lccccc}
\toprule
\rowcolor{gray!30}
Network         & Param. & GFLOPs & Top-1 (\%) & Top-5 (\%) & Throughput (sample/sec)\\
\midrule
ResNet-50       & 25.56M                        & 4.11                            & 76.05       & 92.80        & 2042\\
SRM-ResNet-50\footnote[4]          & 25.62M  & 4.15\scriptsize({$\Delta 0.04$})     & 77.10 & - & 1243\scriptsize({$\Delta 799$})\\
GE-ResNet-50\tablefootnote[4]{Results are from~\cite{nlnbilinear}}           & 31.12M  & 4.14\scriptsize({$\Delta 0.03$})     & 76.80 & - & 1365\scriptsize({$\Delta 677$})\\
SE-ResNet-50           & 28.09M & 4.14\scriptsize({$\Delta 0.03$})   & 76.84       & 93.45       & 1787\scriptsize({$\Delta 255$}) \\
SK-ResNet-50           & 27.49M & 4.47\scriptsize({$\Delta 0.36$})   & 77.56       & 93.62       & 1557\scriptsize({$\Delta 485$}) \\
\rowcolor{red!20}
SB-ResNet-50    & 25.99M & 4.13\scriptsize({$\Delta 0.02$}) & 76.86 & 93.33 & 1836\scriptsize({$\Delta 206$})\\
\midrule
\rowcolor{blue!5}
SB-SE-ResNet-50        & 28.52M &  4.16\scriptsize({$\Delta 0.05$})   & 77.10       & 93.59       & 1626\scriptsize({$\Delta 416$}) \\
\rowcolor{blue!5}
SB-SK-ResNet-50        & 27.94M &  4.49\scriptsize({$\Delta 0.38$})   & 77.93       & 93.54       & 1440\scriptsize({$\Delta 602$}) \\
\bottomrule
\end{tabular}
}
\label{imagenet_results:attention}
\end{table}

\begin{table}
\renewcommand{\arraystretch}{1.22}
\setlength{\tabcolsep}{7pt}
\caption{Experimental result on the comparison with state-of-the-art (SoTA) non-local neural networks. We compare the proposed attention-free spatial bias method with the self-attention based non-local neural networks. Our spatial bias (SB) improve the performance with very few additional resources compared to the SoTA methods. Additionally, due to our lightweight architecture, SB further improve the networks when combining with self-attention based non-local methods.
}
\label{table:Comparision_sota}
{\scriptsize
\begin{tabular}{lccccc}
\toprule
\rowcolor{gray!30}
Network         & Param.                        & GFLOPs                          & Top-1 (\%)  & Top-5 (\%)  & Throughput (sample/sec)\\
ResNet-50       & 25.56M                        & 4.11                            & 76.05       & 92.80        & 2042\\
NLNet-50          & 32.92M & 7.66\scriptsize({$\Delta 3.55$})     & 77.25       & 93.66       & 1353\scriptsize({$\Delta 689$}) \\
GCet-50$_{+all}$  & 28.08M & 4.12\scriptsize({$\Delta 0.01$})     & 76.93       & 93.25       & 1719\scriptsize({$\Delta 323$}) \\
BAT-50          & 30.23M & 5.41\scriptsize({$\Delta 1.30$})     & 77.78       & 94.01       & 1232\scriptsize({$\Delta 810$})  \\
\rowcolor{red!20}
SB-ResNet-50    & 25.99M & 4.13\scriptsize({$\Delta 0.02$})     & 76.86 &  93.33 & 1836\scriptsize({$\Delta 206$})\\ 
\midrule
\rowcolor{blue!5}
SB-NLNet-50       & 33.35M & 7.68\scriptsize({$\Delta 3.57$})     & 77.59       & 93.74       &  1276\scriptsize({$\Delta 766$})    \\
\rowcolor{blue!5}
SB-GCNet-50$_{+all}$&28.51M & 4.14\scriptsize({$\Delta 0.03$})     &  77.00      &  93.27      &  1613\scriptsize({$\Delta 429$})    \\
\rowcolor{blue!5}
SB-BAT-50       & 30.66M & 5.43\scriptsize({$\Delta 1.32$})     &  78.06      & 93.97       &  1153\scriptsize({$\Delta 889$})\\
\bottomrule
\end{tabular}
}
\label{imagenet_results}
\end{table}

\begin{figure*}[t]
\centering
\hspace*{-0.5in}
\includegraphics[width=14.6cm, height=9cm]{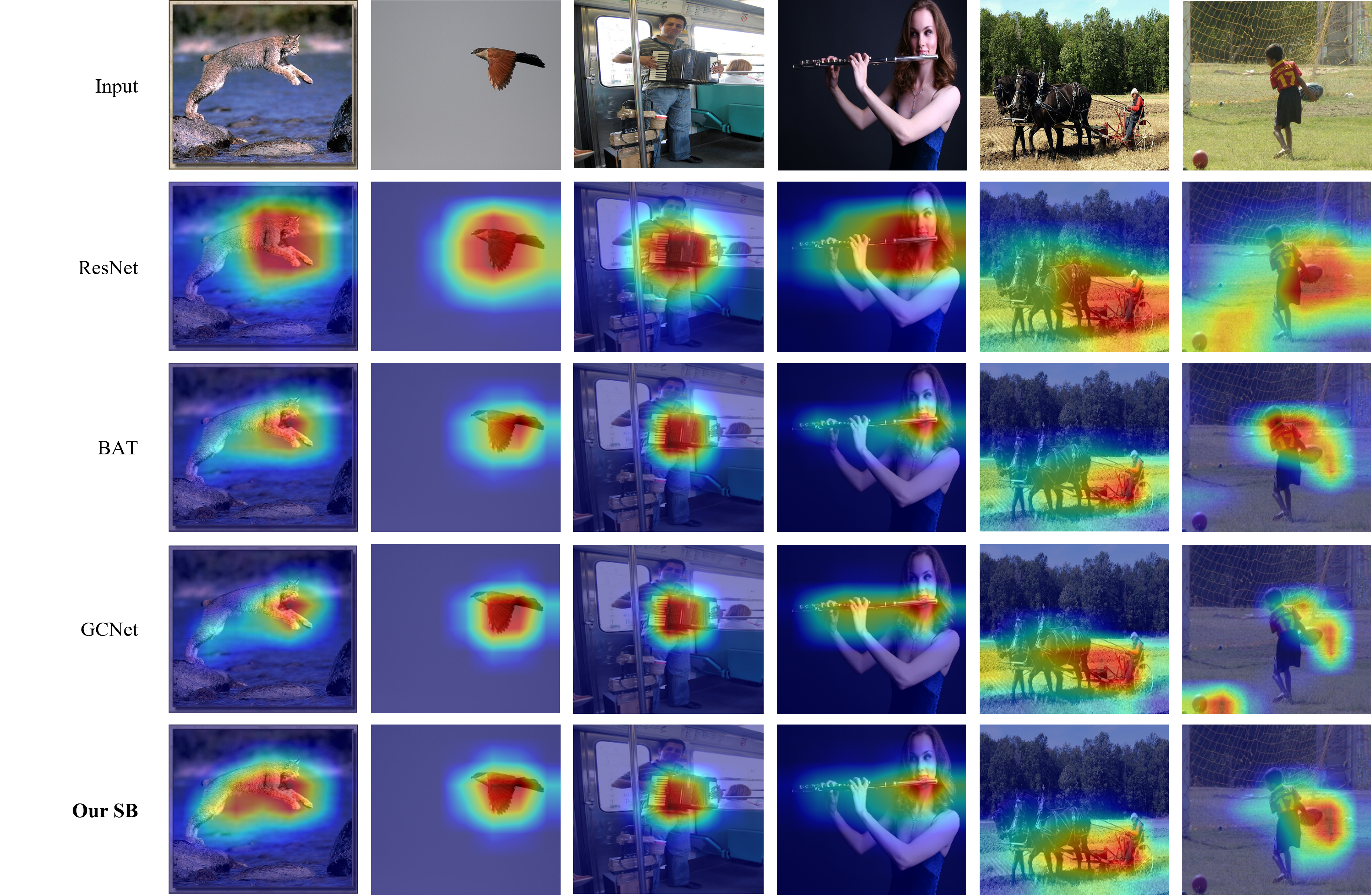} 
\caption{Grad-CAM~\cite{gradcam} visualization of our spatial bias (SB) and non-local methods. 
Our spatial bias focuses more on the discriminant parts of an object.
}
\label{compare_gradcam}
\end{figure*}

\begin{table}[h]
\renewcommand{\arraystretch}{1.2}
\setlength{\tabcolsep}{11pt}
\centering
\caption{Comparison between non-local neural networks and spatial bias on ImageNet-1K.}
\scriptsize{
\begin{tabular}{lccc}
\toprule
\rowcolor{gray!30}
Network & Param. & Top-1 (\%) & {Latency (step/sec)} \\
\midrule
NLNet-50   & 32.9M & 77.2 & $\bf{\Delta689}$ \\
Reduced NLN & 32.9M & 77.0 & $\bf{\Delta630}$ \\
SB (Ours)  & 26.0M & 76.9 & $\bf{\Delta206}$\\
\bottomrule
\end{tabular}
}
\label{new_inference}
\end{table}

In Table~\ref{imagenet_results}, we compare the performance of spatial bias and conventional non-local neural networks~\cite{nln,gcnet,nlnbilinear}.
Our spatial bias need a minimal parameter overhead compared to NLNet~\cite{nln} so that ours is faster than them by 3.3 times. 
Compared with improved version of non-local neural networks (\ie GCNet and BAT)~\cite{gcnet,nlnbilinear}, the computational budget of the spatial bias is much cheaper while achieving comparative performance.
In particular, existing non-local methods (NLNet, GCNet, and BAT) apply the self-attention module in only specific layers due to the heavy design, but the proposed spatial bias can be used to all layers with minimal overhead.
Therefore, our spatial bias is combined with the existing non-local methods in a network to further improve its performance.
We also proceed with the comparison by visualizing the attention map of spatial bias and other non-local neural networks.
As shown in Fig.~\ref{compare_gradcam}, the proposed spatial bias is simple yet straightforward, but the visualization results of our method are comparable to complex self-attention-based non-local neural networks.

\subsection{Compare with Compressed Self-attention}
We conduct further experiments on compressed non-local neural networks. We applied NLNet-50 by compressing the features to $10\times10$ with the same average pooling used in our spatial bias. As shown in Table~\ref{new_inference}, we confirms that the compressed NLNet-50 has little gain in parameters and latency, but its performance deteriorate.

\subsection{OOD distortion Robustness and Shape bias}
In this section, we verify the performance of the proposed spatial bias on the out-of-distribution data set for measuring a statistically different distribution from the training data.
We conduct an OOD distortion robustness experiment on a total of 17 test datasets which have statistically different distributions.
It includes five datasets(sketches~\cite{sketchdataset}, edge-filtered images, silhouettes, texture-shape cue conflict, and stylized images~\cite{cue_conflict}) and 12 test datasets~\cite{12dataset}. 
We compare the OOD robustness of the spatial bias and non-local neural networks using two different metrics (accuracy difference\footnote{Compare machine and human accuracy in various ood tests}, observed consistency\footnote{The fraction of human and models making the same choices (right or wrong)~\cite{ob_consistency}}.
In Table~\ref{ood_results}, the proposed spatial bias is more robust to OOD datasets compared to the conventional non-local neural networks. 
This result prove that the proposed spatial bias works well regardless of the data domain.

\begin{table}
\renewcommand{\arraystretch}{1.2}
\setlength{\tabcolsep}{12pt}
\caption{Experimental results on OOD datasets. 
We compare the OOD robustness using three metrics. Spatial bias shows better OOD robustness compared to non-local neural networks.
} 
\centering
{\scriptsize
\begin{tabular}{l|cc}
\toprule
\rowcolor{gray!30}
Network & Acc diff. $\downarrow$ & Obs.consistency $\uparrow$\\
\midrule
BAT-50 & 0.069 & 0.677  \\
\midrule
\rowcolor{red!20}
SBNet-50 & 0.078 & 0.668  \\
\midrule
GCNet-50 & 0.081 & 0.668  \\
\midrule
NLNet-50 & 0.086 & 0.661  \\
\midrule
ResNet-50 & 0.087 & 0.657  \\
\bottomrule
\end{tabular}
}
\label{ood_results}
\end{table}

\subsection{Object Detection}
In this subsection, we validate the performance of spatial bias on object detection task. 
In this experiment, we use Faster R-CNN~\cite{faster_rcnn} and Cascade R-CNN~\cite{cascade} with FPN~\cite{fpn} using 118k training and 5k verification images from the MS COCO-2017 dataset~\cite{coco}.
We use ResNet-50 as a backbone models previously trained on ImageNet dataset.
By following the standard protocol~\cite{mmdetection}, we use a $1\times$ learning rate schedule with 12 epochs.
We exploit the SGD optimizer with 1e-4 weight decay value and 0.9 momentum, initial learning rate as 0.02.
Networks are trained on two A5000 GPUs with 8 samples per GPU.
We reduce the width of the image to 800 and keep the height below 1,333 pixels.
As shown in Table~\ref{table:compraision_detection} , the networks with our spatial bias improve the performance of detection model for all metrics. 
\begin{table*}[t]
\renewcommand{\arraystretch}{1.2}
\setlength{\tabcolsep}{9.12pt}
\caption{Experimental results on object detection using MS-COCO dataset.}
\centering
{\scriptsize
\begin{tabular}{l|l|c|c|c|c|c|c}
\toprule
\rowcolor{gray!30}
Method & Backbone & $AP$ & $AP_{50}$ & $AP_{75}$ & $AP_{S}$ & $AP_{M}$ & $AP_{L}$\\
\midrule
\multirow{2}{*}{Faster-RCNN} 
 & ResNet-50 & 39.0 & 60.3 & 42.4 & 23.0 & 43.2 & 50.0\\
 &  \cellcolor{red!20}ResNet-50 + ours  & \cellcolor{red!20}{\bf 40.0} & \cellcolor{red!20}{\bf 61.5} & \cellcolor{red!20}{\bf 43.7} & \cellcolor{red!20}{\bf 24.0} & \cellcolor{red!20}{\bf 44.1} & \cellcolor{red!20}{\bf 51.6}\\
\midrule
\multirow{2}{*}{Cascade-RCNN} & ResNet-50 & 41.9 & 60.5 & 45.7 & 24.2 & 45.8 & 54.8\\
                              &  \cellcolor{red!20}ResNet-50 + ours & \cellcolor{red!20}{\bf 42.8} & \cellcolor{red!20}{\bf 61.9} & \cellcolor{red!20}{\bf 46.8} & \cellcolor{red!20}{\bf 25.2} & \cellcolor{red!20}{\bf 46.2} & \cellcolor{red!20}{\bf 55.5} \\
\bottomrule
\end{tabular}
}
\label{table:compraision_detection}
\end{table*}

\begin{table*}[t]
\renewcommand{\arraystretch}{1.2}
\setlength{\tabcolsep}{14pt}
\caption{Experimental result on semantic segmentation using ADE20K. 
}
\centering
\label{table:semantic segmentation}
{\scriptsize
\begin{tabular}{l|l|c|c|c|c}
\toprule
\rowcolor{gray!30}
Method & Backbone & $aAcc$ & $mIoU$ & $mAcc$ & $FPS(img / s)$ \\
\midrule
\multirow{2}{*}{FPN} & ResNet-50 & 77.82 & 38.04 & 48.5 & 40.76\\
                     & \cellcolor{red!20}ResNet-50 + ours & \cellcolor{red!20} {\bf 79.09} & \cellcolor{red!20}{\bf 40.20} & \cellcolor{red!20} {\bf 51.81} & \cellcolor{red!20}32.40\\
\bottomrule
\end{tabular}
}
\end{table*}
\subsection{Semantic Segmentation}
We perform the evaluation of semantic segmentation task using the ADE20k dataset~\cite{ADE20K}.
FPN~\cite{fpn} architecture is utilized for the baseline model to which the spatial bias is applied\footnote[5]{We adopt the implementation of FPN model from~\cite{contributors2020mmsegmentation}.}.
Segmentation networks are trained on two GPUs with 14 samples per GPU for 40K iterations.
Same as the detection networks, we use ResNet-50 backbone model trained on ImageNet with input $512\times512$ input image size.
We employ the AdamW~\cite{adamw} as the optimization algorithm and set the initial learning rate as $2\times10^{-4}$ and a weight decay as $10^{-4}$ with polynomial learning rate decay.
As shown in Table~\ref{table:semantic segmentation}, networks with our spatial bias outperform baseline networks by $+1.27 aAcc$, $+2.16 mIoU$, $+3.31 mAcc$.

\section{Conclusion}

In this paper, we propose the spatial bias that learn global knowledge with fast and lightweight architecture. 
The proposed method adds only a few additional spatial bias channels to a convolutional feature map so that the convolution layer itself learns global knowledge with the self-attention operation.
That is, the spatial bias is be a kind of non-local method that allows convolution to learn long-range dependency.
Spatial bias generates much less parameter, FLOPs, and the throughput overhead than existing non-local methods~\cite{nln,nlnbilinear,ccnet,a2net}.
Our design choice is simple yet straightforward. We assume this is the advantage of being applicable to various network architectures. We argue that the computational cost of the existing non-local neural networks with self-attention operation has increased considerably by using rather complex design choice.
Also, the proposed spatial bias can be used together with existing self-attention based non-local methods. 
We believe that our new approach without self-attention based non local neural networks will inspire future studies.
\\ \\
\bibliographystyle{plainnat}
\bibliography{spatial_bias}

\end{document}